\documentclass[10pt, conference, compsocconf]{llncs}

\usepackage{times}
\usepackage{latexsym}
\usepackage{url}
\usepackage{graphicx}
\usepackage[T1]{fontenc}
\usepackage[utf8]{inputenc}
\usepackage{bm}
\usepackage{amsmath}
\usepackage{amsfonts}
\usepackage{booktabs}
\usepackage{comment}
\usepackage{enumitem}
\usepackage{multirow}
\usepackage{subcaption}
\usepackage[whole]{bxcjkjatype}

\begin{document}

\title{Improving Context-aware Neural Machine Translation with Target-side Context}

\author{Hayahide Yamagishi\inst{1} \and Mamoru Komachi\inst{1}}
\institute{Tokyo Metropolitan University\\
6-6 Asahigaoka, Hino, Tokyo 191-0065, Japan\\
\email{yamagishi-hayahide@ed.tmu.ac.jp, komachi@tmu.ac.jp}
}

\maketitle

\begin{abstract}
  In recent years, several studies on neural machine translation (NMT) have attempted to use document-level context by using a multi-encoder and two attention mechanisms to read the current and previous sentences to incorporate the context of the previous sentences.
  These studies concluded that the target-side context is less useful than the source-side context.
  However, we considered that the reason why the target-side context is less useful lies in the architecture used to model these contexts.

  Therefore, in this study, we investigate how the target-side context can improve context-aware neural machine translation.
  We propose a weight sharing method wherein NMT saves decoder states and calculates an attention vector using the saved states when translating a current sentence.
  Our experiments show that the target-side context is also useful if we plug it into NMT as the decoder state when translating a previous sentence.

\end{abstract}


%

\section{Introduction}

\begin{figure*}[t]
  \begin{center}
    \begin{tabular}{cc}
        \begin{subfigure}{0.5\hsize}
        \begin{center}
          \includegraphics[width=5.5cm]{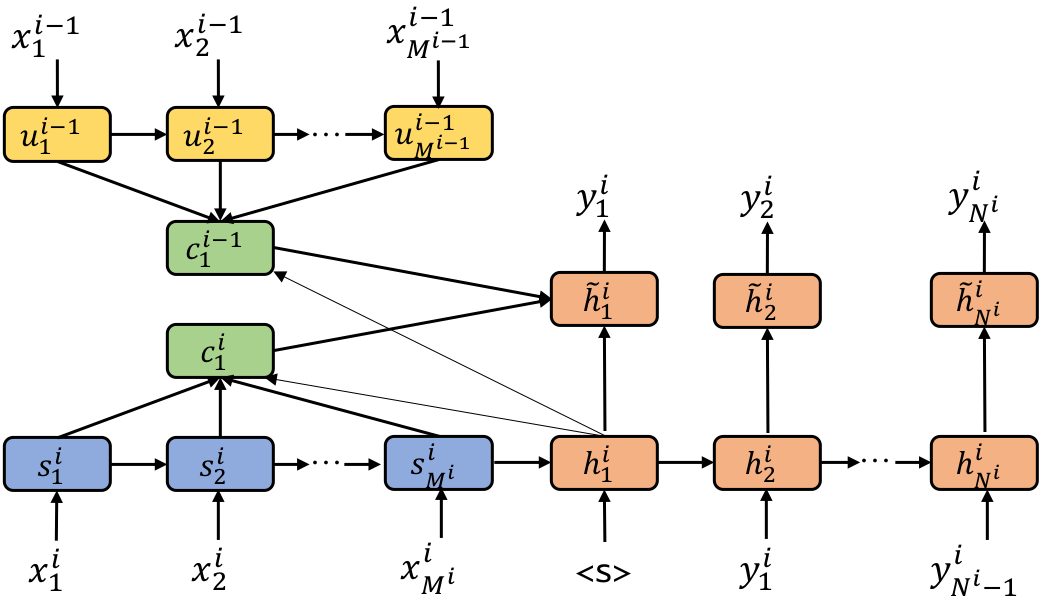} \\
          \caption{Separated source model.}
          \label{sepsrc}
        \end{center}
      \end{subfigure}

      \begin{subfigure}{0.5\hsize}
        \begin{center}
          \includegraphics[width=5.5cm]{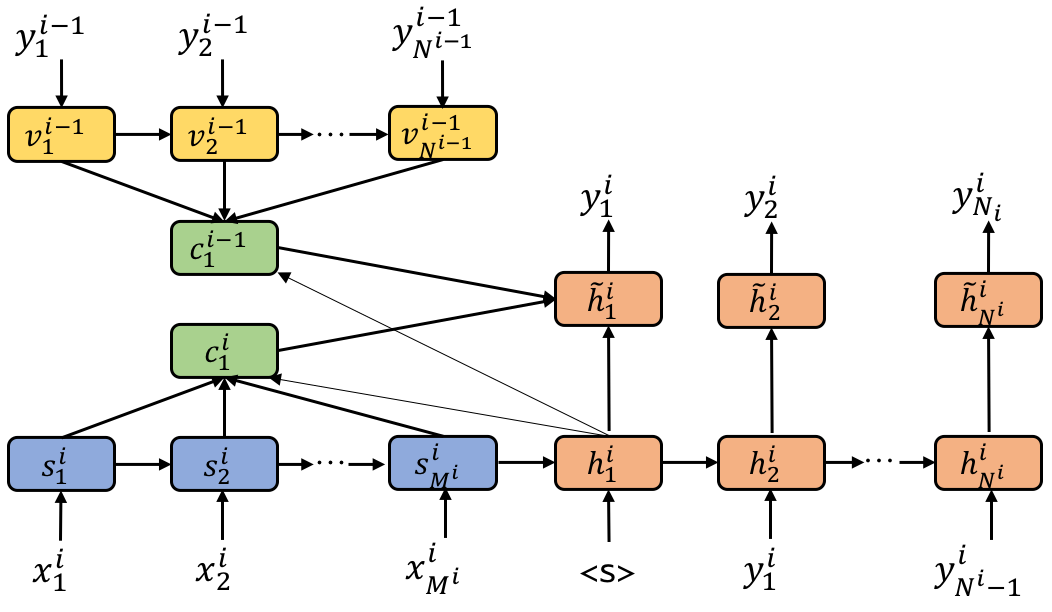} \\
          \caption{Separated target model.}
          \label{septrg}
        \end{center}
      \end{subfigure}

      \\

      \begin{subfigure}{0.5\hsize}
        \begin{center}
          \includegraphics[width=5.5cm]{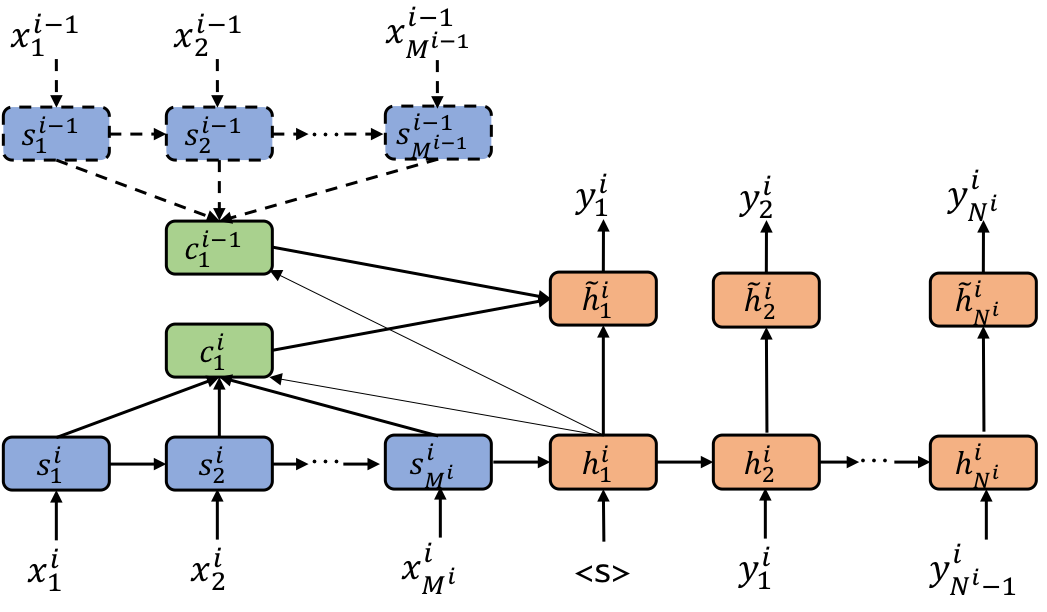} \\
          \caption{Shared source model.}
          \label{shrsrc}
        \end{center}
      \end{subfigure}

      \begin{subfigure}{0.5\hsize}
        \begin{center}
          \includegraphics[width=5.5cm]{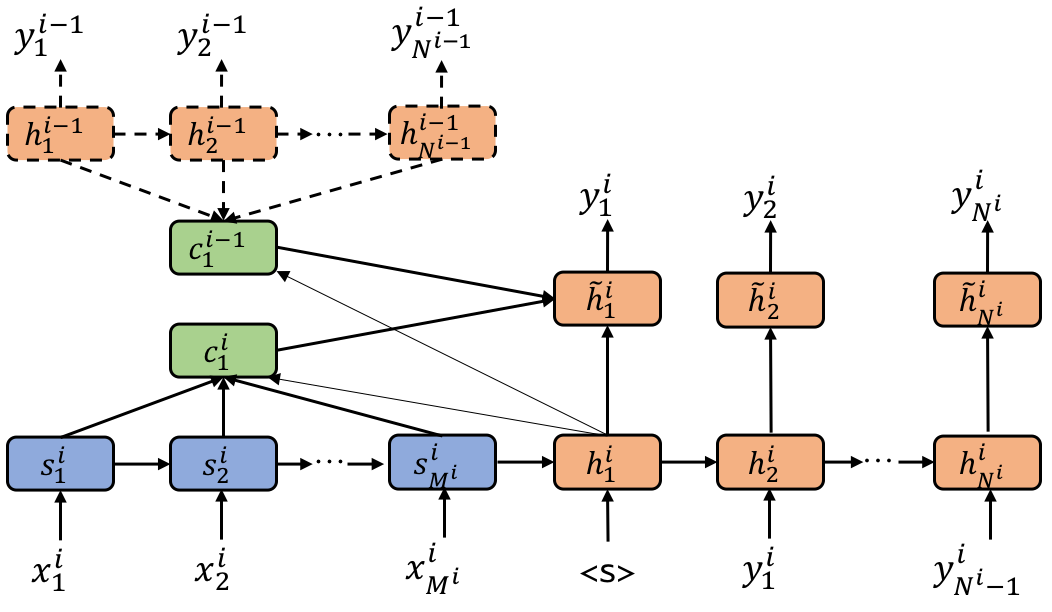} \\
          \caption{Shared target model.}
          \label{shrtrg}
        \end{center}
      \end{subfigure}

    \end{tabular}
    \caption{Proposed methods: dashed line represents the weight sharing with the encoders or decoders.}
    \label{network}
  \end{center}
\end{figure*}

Neural machine translation (NMT; Sutskever et al. \cite{sutskever2014}, Bahdanau et al. \cite{bahdanau2015}, Vaswani et al. \cite{vaswani2017}) has become popular in recent years because it can handle larger contexts compared to conventional machine translation systems.
However, most of the NMTs do not employ document-level contexts due to lack of an efficient mechanism, similar to other machine translation systems.

Recently, a few studies have attempted to expand the notion of a sentence-level context in NMT to that of a document-level context\footnote{Hereinafter, ``document-level context'' is simply referred to as a ``context''.}.
It is reported that the information of one or more previous sentences improves the scores of automatic and human evaluations.

Context-aware NMT systems typically have two encoders: one is for a current sentence and the other is for a previous sentence.
For instance, Bawden et al. \cite{bawden2018} showed that encoding a previous target sentence does not improve the performance in an English--French task even though encoding a previous source sentence works well.
Other studies that utilized a multi-encoder (Jean et al. \cite{jean2017}, Voita et al. \cite{voita2018}, Zhang et al. \cite{zhang2018}) did not use a previous target sentence.
Thus, there are a few works on handling the target-side context.
Moreover, these previous works mainly used language pairs that belonged to the same language family.
In distant language pairs, the information of discourse structures in the target-side document might be useful because distant languages might have different discourse structures.

Therefore, this study investigates how the target-side context can be used in context-aware NMT.
We hypothesize that the source-side contexts should be incorporated into an encoder and the target-side contexts should be incorporated into a decoder.
To validate this hypothesis, we propose a weight sharing method, in which NMT saves the decoder states and calculates an attention vector using the saved states when translating a current sentence.
We find that target-side contexts are also useful if they are inserted into the NMT as the decoder states.
This method can obtain competitive or even better results compared to a baseline model using source-side features.
 
The main findings of this study are as follows:
\begin{itemize}
  \item The target-side context is as important as the source-side context.
  \item The effectiveness of source-side context depends on language pairs.
  \item Weight sharing between current and context states is effective for context-aware NMT.
\end{itemize}


\section{Model Architecture}
\label{Architecture}

Figure \ref{network} presents our methods.
We build context-aware NMT based on the multi-encoder model proposed by Bawden et al. \cite{bawden2018}.
A parallel document $D$ consisting of $L$ sentence pairs, is denoted by $D = (X^1, Y^1), ..., (X^i, Y^i), ..., (X^L, Y^L)$, where $X$ and $Y$ are source and target sentences, respectively.
Each sentence, $X^i$ or $Y^i$, is denoted as $X^i = x_1^i, ..., x_m^i, ..., x_{M^i}^i$ or $Y^i = y_1^i, ..., y_n^i, ..., y_{N^i}^i$, where $x_m^i$ or $y_n^i$ are the tokens, and $M^i$ or $N^i$ are the sentence lengths.
The objective is to maximize the following probabilities:
\begin{eqnarray}
  p(Y^i|X^i, Z^{i-1}) = \prod_{n=1}^{N^i} p(y_n^i|y_{<n}^i, X^i, Z^{i-1})
\end{eqnarray}
where $Z^{i-1}$ represents a previous sentence, $X^{i-1}$ or $Y^{i-1}$, depending on the experimental settings.
Each $p$ is calculated as follows:
\begin{eqnarray}
  p(y_n^i|y_{<n}^i, X^i, Z^{i-1}) &=& \mathrm{softmax} (W_{\mathrm o}\tilde{\bm{h}}_n^i) \\
  \tilde{\bm{h}}_n^i &=& W_{\mathrm h} [\bm{h}_n^i; \bm{c}_n^i; \bm{c}_n^{i-1}] \\
  \bm{c}_n^i &=& \sum_{m=1}^{M^i} \alpha_{n, m}^i \bm{s}_m^i \\
  \alpha_{n, m}^i &=& \mathrm{softmax} (\bm{s}_m^i \cdot \bm{h}_n^i)
\end{eqnarray}
where $\bm{s}_m^i$, $\bm{h}_n^i$, and $\bm{c}_n^i$ represents encoder states, decoder states, and attention, respectively.
$W_{\mathrm o} \in \mathbb{R}^{V \times H}$ and $W_{\mathrm h} \in \mathbb{R}^{H \times 3H}$ represents weights.
We calculate the encoder state $\bm{s}_m^i$ and the decoder state $\bm{h}_n^i$ as follows:
\begin{eqnarray}
  \bm{s}_m^i &=& \mathrm{LSTM}_{enc}(W_{\mathrm x} x_m^i, \bm{s}_{m-1}^i) \\
  \bm{h}_n^i &=& \mathrm{LSTM}_{dec}(W_{\mathrm y} y_n^i, \bm{h}_{n-1}^i) 
\end{eqnarray}
where $W_{\mathrm x} \in \mathbb{R}^{E \times V}$ and $W_{\mathrm y} \in \mathbb{R}^{E \times V}$ represents word embeddings of source- and target sides, respectively.
We use the dot product of encoder states and hidden states as an attention score $\alpha_{n, m}^i$, proposed by Luong et al. \cite{luong2015}.

The multi-encoder model has an additional attention, $\bm{c}_n^{i-1}$, which is for using the information of a previous sentence.
\begin{eqnarray}
  \label{cont_att}
  \bm{c}_n^{i-1} &=& \sum_{t=1}^{|Z^{i-1}|} \beta_{n, t}^{i-1} \bm{z}_t^{i-1} \\
  \beta_{n, t}^{i-1} &=& \mathrm{softmax} (\bm{z}_t^{i-1} \cdot \bm{h}_n^i)
\end{eqnarray}
We experiment using two methods, \textit{separated model} and \textit{shared model}.
The separated model represents the conventional multi-encoder model, and the shared model is our proposed method.
The difference between the two methods is the calculation of $\bm{z}_t^{i-1}$.

\begin{figure*}[t]
  \begin{center}
    \begin{tabular}{cc}
      \includegraphics[width=8cm]{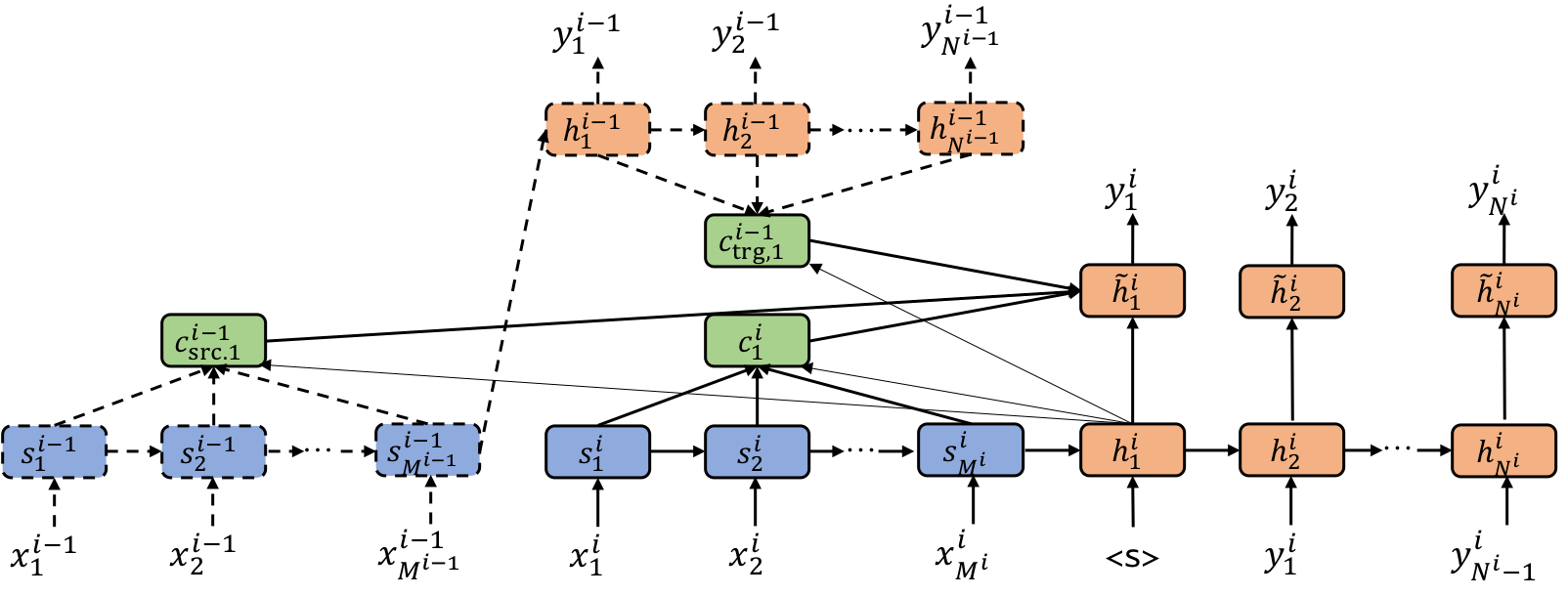} \\
    \end{tabular}
    \caption{Shared mix model.}
    \label{shared_mix_figure}
  \end{center}
\end{figure*}

\subsection{Separated model}
Context-aware NMT saves and encodes a source or target sentence in a context encoder when translating a current sentence.
Previous works on multi-encoder models have an additional encoder, referred to as a context encoder.
Each context encoder $\bm{u}^{i-1}$ or $\bm{v}^{i-1}$ reads a previous source-side or target-side sentence as context, respectively.
\begin{eqnarray}
  \bm{u}_t^{i-1} &=& \mathrm{LSTM}_{\mathrm{src\_enc}}(W_{\mathrm x} x_t^{i-1}, \bm{u}_{t-1}^{i-1}) \\
  \bm{v}_t^{i-1} &=& \mathrm{LSTM}_{\mathrm{trg\_enc}}(W_{\mathrm y} y_t^{i-1}, \bm{v}_{t-1}^{i-1}) 
\end{eqnarray}
We refer to this architecture as a \textit{separated model} in this paper.
In the separated model, the weights of a context encoder are different from those of a current encoder which encodes a current source sentence.
If $\bm{u}_t^{i-1}$ is used as $\bm{z}_t^{i-1}$, we call this model \textit{separated source} model; otherwise, we call this model \textit{separated target} model.

\subsection{Shared model}
A \textit{shared model} saves the hidden states of an encoder or decoder and then calculates $\bm{c}_n^{i-1}$ using these states when translating a current sentence.
The strength of this model is that the target-side context can be incorporated into a decoder instead of an encoder.
Moreover, the shared model does not require much additional parameters and extra computational times because this model simply loads the saved hidden states.
Thus, we can see these models as examples of weight sharing between a current encoder or decoder and a context encoder.
The \textit{shared source} model uses $\bm{s}_t^{i-1}$ as $\bm{z}_t^{i-1}$, and the \textit{shared target} model uses $\bm{h}_t^{i-1}$ as $\bm{z}_t^{i-1}$.

\subsection{Shared mix model}
We propose a \textit{shared mix} model, which incorporates the source- and target-side contexts.
Figure \ref{shared_mix_figure} presents the shared mix model.
The attention vector of the shared mix model $\bm{c}^{i-1}$ is calculated as $\bm{c}^{i-1} = \bm{c}_{source}^{i-1} + \bm{c}_{target}^{i-1}$, where $\bm{c}_{source}^{i-1}$ and $\bm{c}_{target}^{i-1}$ are the context attentions calculated by the equation (\ref{cont_att}).
The reason for calculating the sum of two attention is to arrange the same number of parameters as the other shared models.
Other architectures are the same as the other shared models.


\section{Experiments}
\subsection{Data}

\begin{table*}[t]
  \begin{center}
    \begin{tabular*}{70mm}{@{\extracolsep{\fill}}lrrr} \toprule
      Corpus       & Train  &  Dev  & Test  \\\midrule
      TED De--En    & 203,998 &   888 & 1,305 \\
      TED Zh--En    & 226,196 &   879 & 1,297 \\
      TED Ja--En    & 194,170 &   871 & 1,285 \\
      Recipe Ja--En & 108,990 & 3,303 & 2,804 \\ \bottomrule
    \end{tabular*}
    \caption{Number of sentences in each dataset.}
    \label{corpus}
  \end{center}
\end{table*}

\begin{table*}[t]
  \begin{center}
    \begin{tabular}{lrrrrrr} \toprule
      \multirow{2}{*}{\centering Experiment} & \multirow{2}{*}{\centering Baseline} & \multicolumn{2}{c}{Separated} & \multicolumn{3}{c}{Shared} \\ 
       & & Source & Target & Source & Target & Mix \\ \midrule
      TED De--En & 26.55 & $26.29 \pm .37$ & $26.52 \pm .12$ & $^*27.20 \pm .11$ & $^*\textbf{27.34} \pm .11$ & $27.18 \pm .21$\\ 
      TED En--De & 21.26 & $21.04 \pm .64$ & $20.77 \pm .10$ & $21.63 \pm .27$ & $\textbf{21.83} \pm .30$ & $21.50 \pm .29$ \\ 
      TED Zh--En & 12.54 & $12.52 \pm .33$ & $12.63 \pm .24$ & $^*13.36 \pm .41$ & $^*\textbf{13.52} \pm .10$ & $^*13.23 \pm .09$ \\ 
      TED En--Zh &  8.97 & $8.94 \pm .11$  & $8.71 \pm .06$  & $9.45 \pm .22$  & $^*\textbf{9.58} \pm .13$ & $9.42 \pm .19$ \\ 
      TED Ja--En &  5.84 & $^*6.64 \pm .26$ & $^*6.37 \pm .12$ & $^*6.95 \pm .07$   & $^*\textbf{6.96} \pm .18$  & $^*6.81 \pm .16$ \\ 
      TED En--Ja &  8.40 &  $8.58 \pm .12$ &  $8.26 \pm .00$ & $8.51 \pm .31$ &  $8.59 \pm .08$ &  $\textbf{8.66} \pm .14$ \\
      Recipe Ja--En & 25.34 & $^*26.51 \pm .09$ & $^*26.69 \pm .15$ & $^*26.90 \pm .17$ & $^*\textbf{26.92} \pm .10$ & $^*26.78 \pm .11$ \\
      Recipe En--Ja & 20.81 & $^*21.87 \pm .12$ & $^*21.45 \pm .14$ & $^*\textbf{22.02} \pm .20$ & $^*21.97 \pm .09$ & $^*21.81 \pm .15$ \\ \bottomrule
    \end{tabular}
  \caption{BLEU scores of our context-aware NMT in each language pair. Each score is the average of three runs. ``$*$'' represents the statistically significant results against the baseline at $p < 0.05$ in all the runs.}
  \label{bleu}
  \end{center}
\end{table*}

We mainly use the IWSLT2017 German--English, Chinese--English, and Japanese--English datasets from TED \cite{ted2012} for experiments.
We consider each talk of TED as a document, which includes sentences that cannot be translated using only sentence-level information.
Japanese and Chinese sentences are segmented by the MeCab\footnote{http://taku910.github.io/mecab/} (dictionary: IPADic 2.7.0) and jieba\footnote{https://github.com/fxsjy/jieba}, respectively.
English and German sentences are segmented by \texttt{tokenizer.perl} included in Moses\footnote{http://www.statmt.org/moses/}.
The documents that include sentences consisting of more than 100 words are eliminated from the training corpus.
We evaluate our methods on the 2014 test set.
The statistics of preprocessed corpora are shown in Table \ref{corpus}.
Byte pair encoding \cite{sennrich2016} is used separately for source and target languages for subword segmentation.
The number of merge operations is 32,000.

Moreover, we use the Recipe Corpus\footnote{http://lotus.kuee.kyoto-u.ac.jp/WAT/recipe-corpus/}, which consists of Japanese--English user-posted recipes, to investigate the influences in the different domains.
The procedures of data preprocessing are the same as those for the TED corpus, except for the number of merge operations (8,000).


\subsection{Settings}
\label{Settings}

The baseline system of this experiment is our implementation of RNN-based NMT.
The encoder is two-layer bi-LSTM, and the decoder is two-layer uni-LSTM.
The dimensions of hidden states and embeddings are set to be 512.
We use dropout with $p = 0.2$.
The optimizer is AdaGrad with initial learning rate $= 0.01$.
Each batch consists of up to 128 documents.
These settings are the same in the baseline and all context-aware models.
Dot global attention is used for calculating context attention $\bm{c}^{i-1}$.
We set $\bm{c}^{0} = \bf{0}$ because the first sentences in documents do not have any previous contexts.

The context-aware models are pretrained with the baseline system.
Each model is trained for 30 epochs; then, the best model is selected with a development set.
The results are evaluated using BLEU \cite{papineni2002}.
We calculate the statistical significance between the baseline and our methods by the bootstrap resampling toolkit in Travatar \cite{neubig2013}.
Experiments are performed three times with different random seeds.


\subsection{Results}

Table \ref{bleu} shows the results.
The shared target model improves the performances in all language pairs.
In the experiments on several language pairs, the separated target model used in Bawden et al. \cite{bawden2018} also improves performances compared to the baseline.
However, improvement is less compared to the shared target model.
Therefore, these results show that the target-side context should be introduced from a decoder.


\section{Discussion}

\begin{table*}[t]
  \begin{center}
  \renewcommand{\arraystretch}{1.19}
  \begin{tabular}{|c|p{10cm}|} \hline
      Experiment & Sentences \\\hline\hline
      \multirow{4}{*}{\centering Input} & わかめ は よく 洗っ て 塩 を 落とし 、 10 分 ほど 水 に 浸け て おい て から ざく 切り に する 。 \textbf{長 ねぎ} は 小口切り に する 。\\ \cline{2-2}
      & 熱し た 鍋 に ごま油 を ひき 、 わかめ と \textbf{長 ねぎ} を 入れ て 30 秒 ほど 軽く 炒める 。\\ \hline

      \multirow{4}{*}{\centering Reference} & wash the wakame well to remove the salt , put into a bowl of water for 10 minutes and drain . cut into large pieces . slice the \textbf{Japanese leek} .\\ \cline{2-2}
      & heat a pan and pour the sesame oil . stir fry the wakame and \textbf{leek} for 30 seconds .\\ \hline

      \multirow{4}{*}{\centering Baseline} & wash the wakame seaweed well and remove the salt . soak in water for 10 minutes , then roughly chop . cut the \textbf{Japanese leek} into small pieces .\\ \cline{2-2}
      & heat sesame oil in a heated pot , add the wakame and \textbf{leek} , and lightly saut\'e for about 30 seconds .\\ \hline

      \multirow{4}{*}{\centering Separated Source} & wash the wakame well , remove the salt , soak in water for about 10 minutes , then roughly chop . cut the \textbf{Japanese leek} into small pieces .\\ \cline{2-2}
      & heat sesame oil in a heated pot and add the wakame and \textbf{leek} . stir-fry for about 30 seconds .\\ \hline

      \multirow{4}{*}{\centering Shared Source} & wash the wakame well, remove the salt , soak in water for about 10 minutes , then roughly chop . cut the \textbf{Japanese leek} into small pieces .\\ \cline{2-2}
      & heat sesame oil in a heated pot and add the wakame and \textbf{leek} . stir-fry for about 30 seconds .\\ \hline

      \multirow{4}{*}{\centering Separated Target} & wash the wakame well , soak in water for about 10 minutes . cut into small pieces. cut the \textbf{Japanese leek} into small pieces .\\ \cline{2-2}
      & heat the sesame oil in a frying pan , add the wakame and \textbf{leek} , and stir-fry for about 30 seconds .\\ \hline

      \multirow{4}{*}{\centering Shared Target} & wash the wakame well , remove the salt , soak in water for about 10 minutes , then roughly chop . chop the \textbf{Japanese leek} into small pieces .\\ \cline{2-2}
      & heat sesame oil in a heated pan , add the wakame and \textbf{Japanese leek} , and lightly stir-fry for about 30 seconds .\\ \hline

      \multirow{4}{*}{\centering Shared Mix} & wash the wakame well , remove the salt , soak in water for about 10 minutes , then roughly chop . chop the \textbf{Japanese leek} into small pieces .\\ \cline{2-2}
      & heat sesame oil in a heated pan , add the wakame and \textbf{Japanese leek} , and stir-fry for about 30 seconds .\\ \hline

  \end{tabular}
  \caption{The output examples in Recipe Japanese--English experiments. Each upper sentence represents a previous sentence, and each lower sentence represents a current sentence. Each sequence may comprise several sentences because each sentence in Recipe corpus corresponds to ``one step'' of cooking.}
  \label{extable}
  \end{center}
\end{table*}

\begin{figure*}[t]
  \begin{center}
      \begin{tabular}{cc}
      \begin{subfigure}{0.5\hsize}
          \begin{center}
          \includegraphics[width=5.5cm]{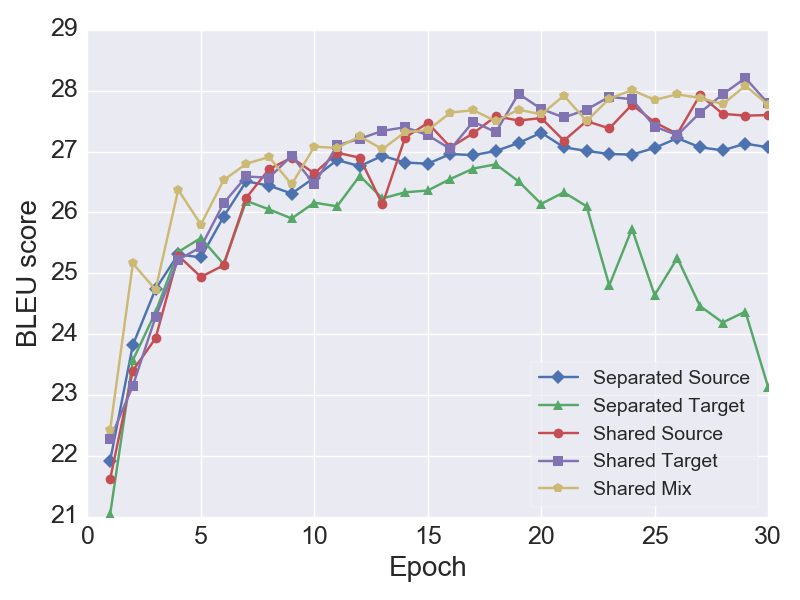} \\
          \caption{TED German--English}
          \end{center}
      \end{subfigure}
  
      \begin{subfigure}{0.5\hsize}
          \begin{center}
          \includegraphics[width=5.5cm]{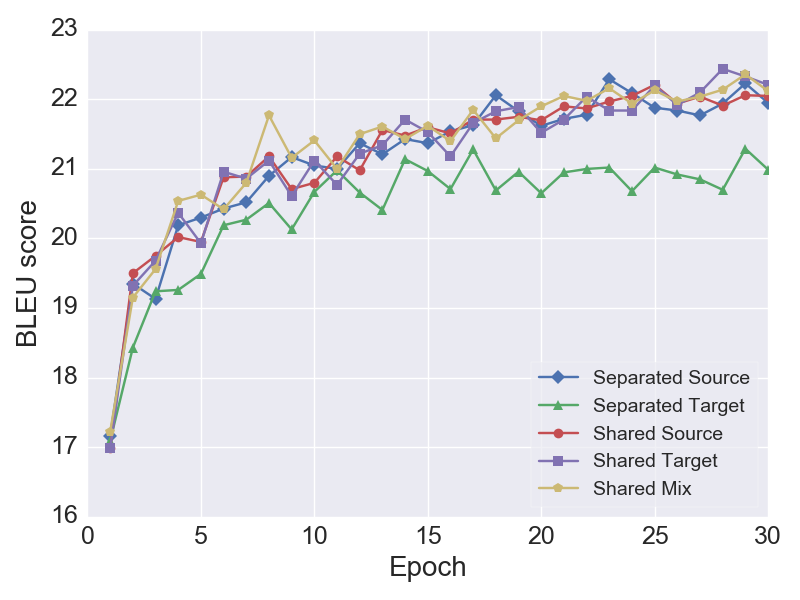} \\
          \caption{TED English--German}
          \end{center}
      \end{subfigure}
  
      \\

      \begin{subfigure}{0.5\hsize}
          \begin{center}
              \includegraphics[width=5.5cm]{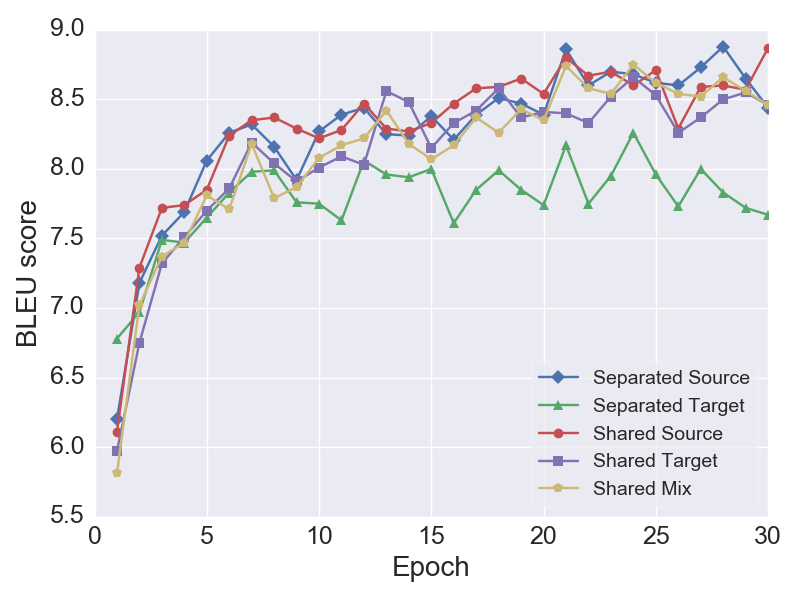} \\
              \caption{TED Chinese--English}
          \end{center}
      \end{subfigure}

      \begin{subfigure}{0.5\hsize}
          \begin{center}
              \includegraphics[width=5.5cm]{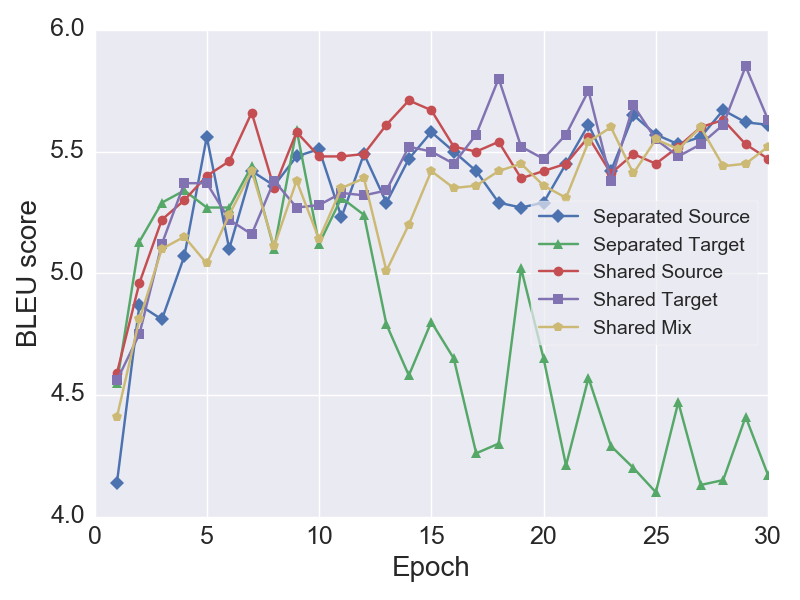} \\
              \caption{TED English--Chinese}
          \end{center}
      \end{subfigure}

      \\

      \begin{subfigure}{0.5\hsize}
          \begin{center}
              \includegraphics[width=5.5cm]{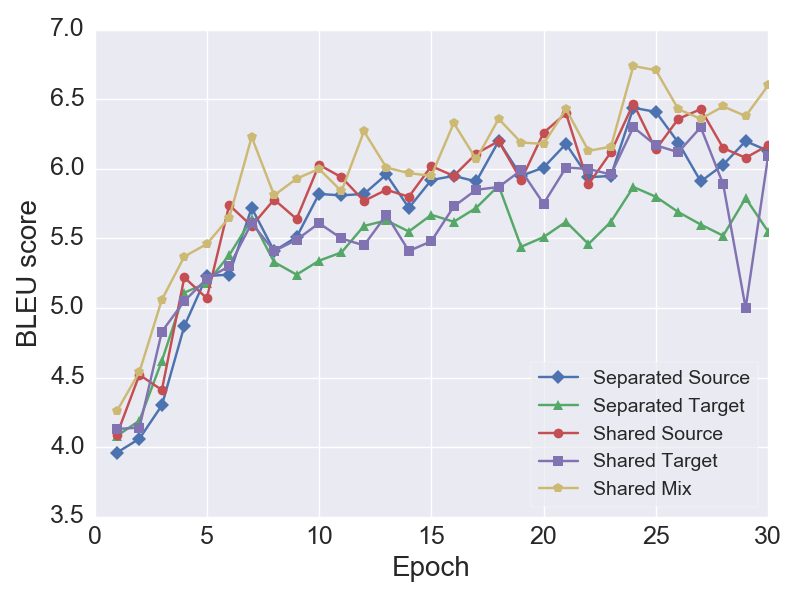} \\
              \caption{TED Japanese--English}
          \end{center}
      \end{subfigure}

      \begin{subfigure}{0.5\hsize}
          \begin{center}
              \includegraphics[width=5.5cm]{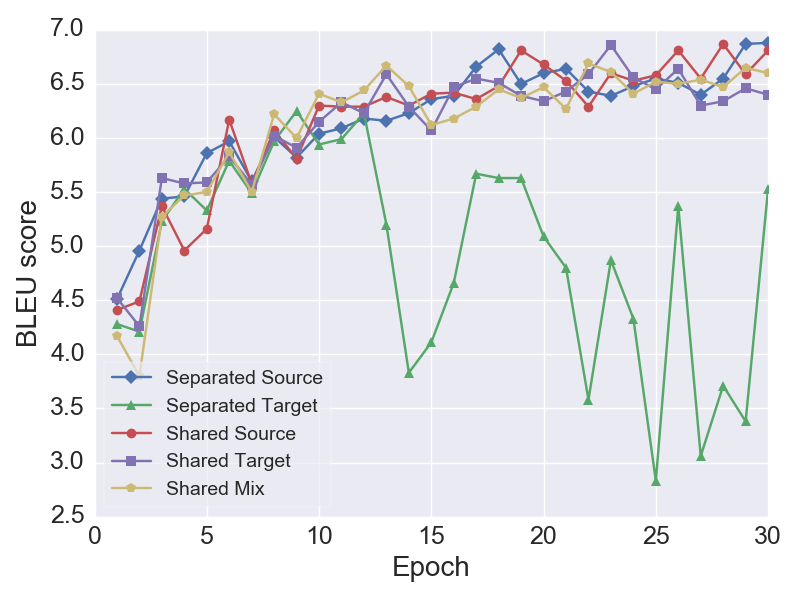} \\
              \caption{TED English--Japanese}
          \end{center}
      \end{subfigure}

      \\

      \begin{subfigure}{0.5\hsize}
          \begin{center}
              \includegraphics[width=5.5cm]{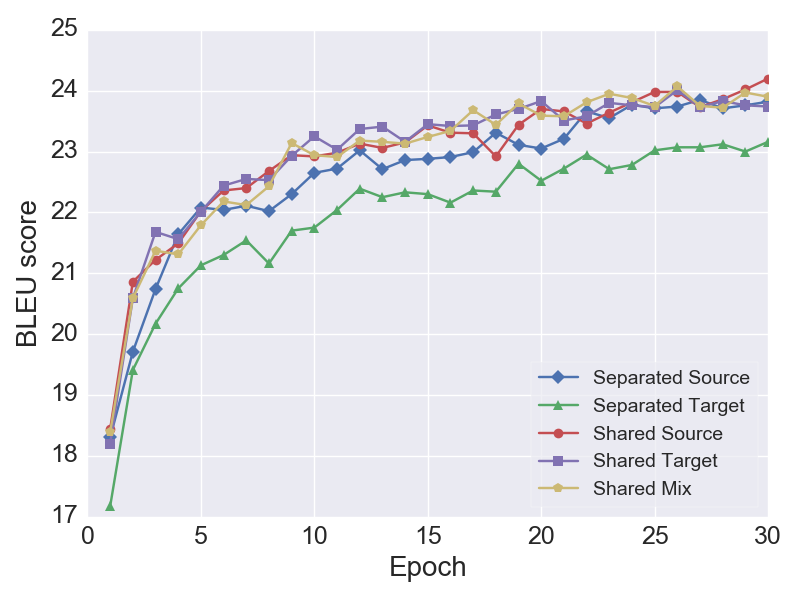} \\
              \caption{Recipe Japanese--English}
          \end{center}
      \end{subfigure}

      \begin{subfigure}{0.5\hsize}
          \begin{center}
              \includegraphics[width=5.5cm]{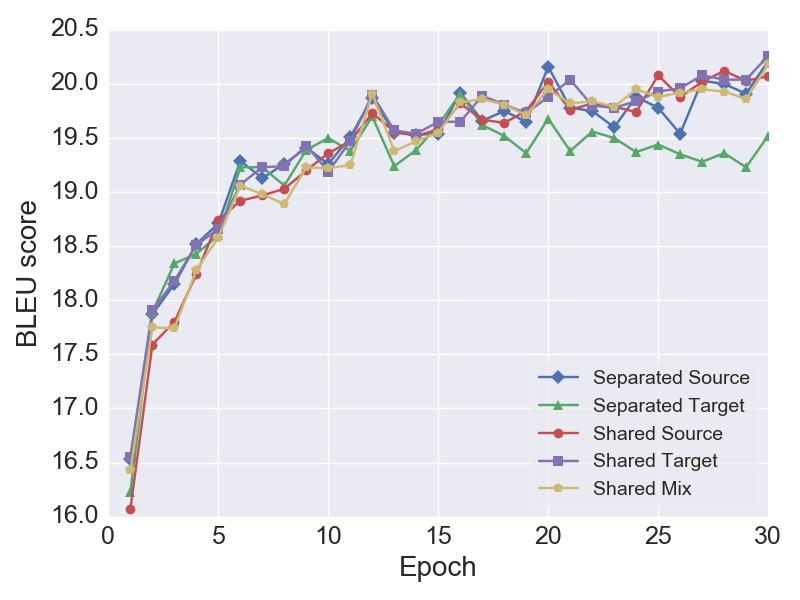} \\
              \caption{Recipe English--Japanese}
          \end{center}
      \end{subfigure}

      \end{tabular}
      \caption{The graph of BLEU scores using each development set. BLEU score is calculated at the end of each epoch.}
      \label{dev_graph_all}
  \end{center}
\end{figure*}

\subsection{Weight sharing}
We expected that there would be no differences between the results of the shared source and separated source models because both models can introduce source-side context into the encoder.
However, the results obtained for the language pairs used in this study show that the shared source model also improves the BLEU scores with fewer parameters.
Dabre et al. \cite{dabre2018} found that translation performances could be boosted even if the weights of stacked layers were shared.
Our shared models can be seen as an instance of weight sharing for stacking sentence-level RNNs in chronological order.
Shared models can also be seen as an instance of multitask learning that shares the same weights for encoder--decoders of neighboring sentences such as skip-thought \cite{kiros2015}.
Thus, it is possible that weight sharing leads to a more efficient model space by regularization, rather than by learning discourse structures.

\subsection{Language dependency}
The tendencies of the scores vary depending on language pairs.
The result of the TED English--German task shows that the source-side context decreases the performance.
M{\"u}ller et al. \cite{muller2018} obtained similar results in other datasets using the concatenation method proposed in Tiedemann et al. \cite{tiedemann2017}.
However, in the Japanese--English and English--Japanese tasks, the importance of the source-side context is equivalent to that of the target-side one.
The reason is that Japanese requires contexts more than English because Japanese is a pro-drop language, which allows for the omission of agents and object arguments when they are pragmatically or syntactically inferable.
Comparing the result of the TED and Recipe corpora, the difference of corpus domains does not affect such tendencies. 
In the Chinese--English task, where they have more similar word order, the importance of target-side context is equivalent or even better compared to that of the source-side one.
Therefore, these results imply that the necessity of the source-side context depends on language pairs, while the target-side context is generally important.

The shared mix model obtains competitive results compared to the shared source model, in most of the language pairs.
Therefore, either of contexts helps the improvement without both side information if we choose the source- or target-side context depending on the language pairs.

\subsection{Output examples}
We analyze the output examples in terms of the phrase coherence.
We select the Recipe Japanese--English task because Japanese is a pro-drop language that needs context due to many omissions but it is difficult to draw any definitive conclusions on the TED Japanese--English task as the BLEU score is too low to analyze.
Table \ref{extable} shows the examples.
When the model translates the previous sentence, this model does not use the context information because this is the first sentence of a document.
The examples written in each lower row are the result using the information of the upper sentence as a context.

Looking at the result of the baseline, ``長 ねぎ'' (naga negi, \textit{Japanese leek}) is translated into ``Japanese leek'' in the previous sentence, even though this is translated into ``leek'' in the current sentence.
This phenomenon can be commonly seen in the results of separated models.
If we independently evaluate these sentences, these sentences will be rated with high fluency and adequacy.
BLEU scores are also high because reference sentences also follow this translation. 
However, these sentences have low coherence because the same noun phrase in the Japanese sentence is translated into different phrases.

On the contrary, ``長 ねぎ'' is translated into ``Japanese leek'' in both sentences in the experiments of shared target model and shared mix model, which use target-side context.
Our models improve phrase coherence using weight sharing.

\subsection{Convergence of training}
Figure \ref{dev_graph_all} plots the BLEU scores on development sets.
Shared models and separated source model seem to be stable.
However, the separated target model is unstable and does not lead to an improvement.
This is due to the exposure bias problem \cite{ranzato2015} in the context encoder as well as the decoder.
At the test phase, the separated target model has to read the low-quality sentence with well-trained encoder if the learning speed of encoding is faster than that of decoding.
Thus, the separated target model should fill the gap between the learning speed of the context encoder and decoder.


\section{Related Works}
Wang et al. \cite{wang2017}, Maruf et al. \cite{maruf2018}, and Tu et al. \cite{tu2018} incorporated the information of previous sentences by using a hierarchical encoder, a memory network and cache mechanism respectively.
Although they used several sentences as contexts, the former two works found that the information of distant sentences in a document does not improve translation quality.
Our investigation is focused on a previous sentence.

Tiedemann et al. \cite{tiedemann2017} used the concatenation of a previous sentence and a current sentence as an input or output sentence to incorporate source-side and target-side contexts in conventional NMT.
M{\"u}ller et al. \cite{muller2018} evaluated the performance of existing context-aware NMT in the English--German task in terms of pronoun translation.
They concluded that generating concatenated sentence is more effective than inputting concatenated sentence.
Our results of the shared target model support their results.

Voita et al. \cite{voita2018} and Zhang et al. \cite{zhang2018} proposed Transformer-based context-aware NMT.
The former suggested that self-attention solves anaphora resolution.
The latter performed fine-tuning with small document-level data to adapt a single-sentence NMT trained with large data to context-aware NMT.
However, they did not investigate the influence of the target-side contexts.


\section{Conclusion}
We reported how context-aware neural machine translation effectively employs target-side contexts.
We proposed a weight sharing to model the target-side context in a decoder.
This method achieves high performances in several language pairs, even though it does not require much additional parameters.
In the future, we will analyze whether this model can handle longer contexts.

\clearpage

\bibliographystyle{IEEEtran}
\bibliography{reference}

\end{document}